\documentclass[11pt,a4paper]{article}
\usepackage[hyperref]{emnlp2020}
\usepackage{times}
\usepackage{latexsym}

\usepackage{microtype}

\usepackage{graphicx}
\usepackage{amsmath}
\usepackage{amsthm}
\usepackage{booktabs}
\usepackage{algorithm}
\usepackage{algorithmic}
\usepackage{array}
\usepackage{float}
\usepackage{booktabs}

\usepackage{graphicx}
\usepackage{subfigure}
\usepackage{mathrsfs}
\usepackage{amsfonts}

\usepackage{paralist}

\usepackage{xurl}

\aclfinalcopy 
\def\aclpaperid{1798} 

\title{Multi-Unit Transformers for Neural Machine Translation}

\author{
Jianhao Yan\qquad Fandong Meng \qquad Jie Zhou\\
Pattern Recognition Center, WeChat AI, Tencent Inc, Beijing, China \\
\{elliottyan,fandongmeng,withtomzhou\}@tencent.com
}

\date{}

\begin{document}

\setlength{\abovedisplayskip}{4pt}
\setlength{\belowdisplayskip}{4pt}

\maketitle
\begin{abstract}
    Transformer models \cite{vaswani2017attention} achieve remarkable success in Neural Machine Translation. 
    Many efforts have been devoted to deepening the Transformer by stacking several units (i.e., a combination of Multihead Attentions and FFN) in a cascade, 
    while the investigation over multiple parallel units draws little attention.
    In this paper, we propose the \textbf{M}ulti-\textbf{U}nit \textbf{T}ransform\textbf{E}rs (\textbf{MUTE}), which aim to promote the expressiveness of the Transformer by introducing diverse and complementary units.
    Specifically, we use several parallel units and show that modeling with multiple units improves model performance and introduces diversity. 
    Further, to better leverage the advantage of the multi-unit setting, we design biased module and sequential dependency that guide and encourage complementariness among different units. 
    Experimental results on three machine translation tasks, the NIST Chinese-to-English, WMT'14 English-to-German and WMT'18 Chinese-to-English, show that the MUTE models significantly outperform the Transformer-Base, by up to +1.52, +1.90 and +1.10 BLEU points, with only a mild drop in inference speed (about 3.1\%). 
    In addition, our methods also surpass the Transformer-Big model, with only 54\% of its parameters. These results demonstrate the effectiveness of the MUTE, as well as its efficiency in both the inference process and parameter usage. \footnote{Code is available at \url{https://github.com/ElliottYan/Multi\_Unit\_Transformer}}
\end{abstract}

\section{Introduction}
Transformer based models \cite{vaswani2017attention} have been proven to be very effective in building the state-of-the-art Neural Machine Translation (NMT) systems via neural networks and attention mechanism \cite{sutskever2014sequence,bahdanau2014neural}. Following the standard \emph{Sequence-to-Sequence} architecture, Transformer models consist of two essential components, namely the encoder and decoder, which rely on stacking several identical layers, i.e., multihead attentions and position-wise feed-forward network. 

Multihead attentions and position-wise feed-forward network, together as a basic unit, 
plays an essential role in the success of Transformer models. 
Some researchers \cite{bapna2018training,wang2019learning} propose to improve the model capacity by stacking this basic unit many times, i.e., deep Transformers, and achieve promising results.
Nevertheless, as an orthogonal direction, investigation over multiple parallel units draws little attention. 

Compared with single unit models, multiple parallel unit layout is more expressive to capture complex information flow \cite{tao2018get,meng2019multi,li2018multi,li2019neuron} in two aspects.
First, this multiple-unit layout boosts the model by its varied feature space composition and different attentions over inputs. With this diversity, multi-unit models advance in expressiveness. 
Second, for the multi-unit setting, one unit could mitigate the deficiency of other units and compose a more expressive network, in a complementary way.

In this paper, we propose the \textbf{M}ulti-\textbf{U}nit \textbf{T}ransform\textbf{E}rs (\textbf{MUTE}), which aim to promote the expressiveness of transformer models by introducing diverse and complementary parallel units.
Merely combining multiple identical units in parallel improves model capability and diversity by its varied feature compositions.
Furthermore, inspired by the well-studied bagging \cite{breiman1996bagging} and gradient boosting algorithms \cite{friedman2001greedy} in the machine learning field, we design biased units with a sequential dependency to further boost model performance. 
Specifically, with the help of a module named bias module, we apply different kinds of noises to form biased inputs for corresponding units. By doing so, we explicitly establish information gaps among units and guide them to learn from each other. 
Moreover, to better leverage the power of \emph{complementariness}, we introduce sequential ordering into the multi-unit setting,
and force each unit to learn the residual of its preceding accumulation.

We evaluate our methods on three widely used Neural Machine Translation datasets, NIST Chinese-English,  WMT'14 English-German and WMT'18 Chinese-English.
Experimental results show that our multi-unit model yields an improvement of +1.52, +1.90 and +1.10 BLEU points, over the baseline model (Transformer-Base) for three tasks with different sizes, respectively. 
Our model even outperforms the Transformer-Big on the WMT'14 English-German by 0.7 BLEU points with only 54\% of parameters. 
Moreover, as an interesting side effect, our model only introduces mild inference speed decrease (about 3.1\%) compared with the Transformer-Base model, and is faster than the Transformer-Big model.

The contributions of this paper are threefold:
\begin{itemize}
\vspace{-5pt}
\setlength{\topsep}{1pt}
\setlength{\itemsep}{1pt}
\setlength{\parsep}{1pt}
\setlength{\parskip}{1pt}
\item We propose the \textbf{M}ulti-\textbf{U}nit \textbf{T}ransform\textbf{E}rs (\textbf{MUTE}), to promote the expressiveness of Transformer models by introducing diverse and complementary parallel units.
\item Aside from learning with identical units, we extend the \textbf{MUTE} by introducing bias module and sequential ordering to further model the diversity and complementariness among different units.
\item Experimental results show that our models substantially surpass baseline models in three NMT datasets, NIST Chinese-English, WMT'14 English-German and WMT'18 Chinese-English. In addition, our models also show high efficiency in both the inference speed and parameter usage, compared with Transformer baselines.
\vspace{-5pt}
\end{itemize}

\section{Transformer Architecture}
The Transformer Architecture \cite{vaswani2017attention} for Neural Machine Translation (NMT) generally adopts the standard encoder-decoder paradigm. In contrast to RNN architectures, the Transformer stacks several identical self-attention based layers instead of recurrent units for better parallelization. 
Specifically, given an input sequence $X=\{x_1, x_2, \cdots, x_n\}$ in source language (e.g., English), the model is asked to predict its corresponding translation $Y=\{y_1, y_2, \cdots, y_m\}$ in target language (e.g., German).

\vspace{-5pt}

\paragraph{Encoder.} Digging into the details of the model, a Transformer encoder consists of $N_e$ stacked layers, where each layer consists of two sub-layers, a multihead self-attention sub-layer and a position-wise feed-forward network (FFN) sub-layer. 

\vspace{-15pt}
\begin{gather}
    s^k = \textup{SelfAttn} (X^k) + X^k, \\
    F^e(X^{k}) = s^k + \textup{FFN}(s^k),
    \vspace{-15pt}
\end{gather}
where $X^k \in \mathbb{R}^{n \times d}$ and $F^e(X^{k}) \in \mathbb{R}^{n \times d}$ denote the inputs and outputs of the $k$-th encoder layer, respectively, and $d$ is the hidden dimension.

\vspace{-5pt}

\paragraph{Decoder.} The decoder follows a similar architecture, with an additional multihead cross-attention sub-layer for each of $N_d$ decoder layers.

\vspace{-15pt}
\begin{gather}
    s^k = \textup{SelfAttn} (Y^k) + Y^k, \\
    c^k = \textup{CrossAttn} (s^k, F^e(X^{N_e})) + s^k, \\
    F^d(Y^{k}) = c^k + \textup{FFN}(c^k),
    \vspace{-15pt}
\end{gather}
where $Y^k\in \mathbb{R}^{m \times d}$ and $F^d(Y^k) \in \mathbb{R}^{m \times d}$ represent the inputs and outputs of $k$-th decoder layer. 

Here, we omit layer norms among sub-layers for simplicity. 
We take the bundle of cascading sub-components (i.e., attention modules and FFN) as a unit, and refer to the original Transformer and its variants with 
such cascade units as \emph{Single-Unit Transformer}.
For ease of reading, we refer to a single unit of encoder and decoder as $F^e$ and $F^d$ in the following sections.

Then, the probability $P(Y|X)$ is produced with another Softmax layer on top of decoder outputs,
\begin{gather}
    P(Y|X) = \textup{Softmax} (W_s \cdot F^d(Y^{N_d}) + b),
\end{gather}
where $W_s \in \mathbb{R}^{d \times |V|}$ and $b \in \mathbb{R}^{|V|}$ are learnable parameters, and $|V|$ denotes the size of target vocabulary. Then, a cross-entropy objective is computed by,
\begin{gather}
    \label{eq:loss}
    \mathcal{L}_{CE} = \sum_{t\in(1, m)} Y_t \log P(Y_t | X),
\end{gather}
where $t$ represents the $t$-th step for decoding phase.

\begin{figure*}
    \centering
    \includegraphics[width=0.92\textwidth]{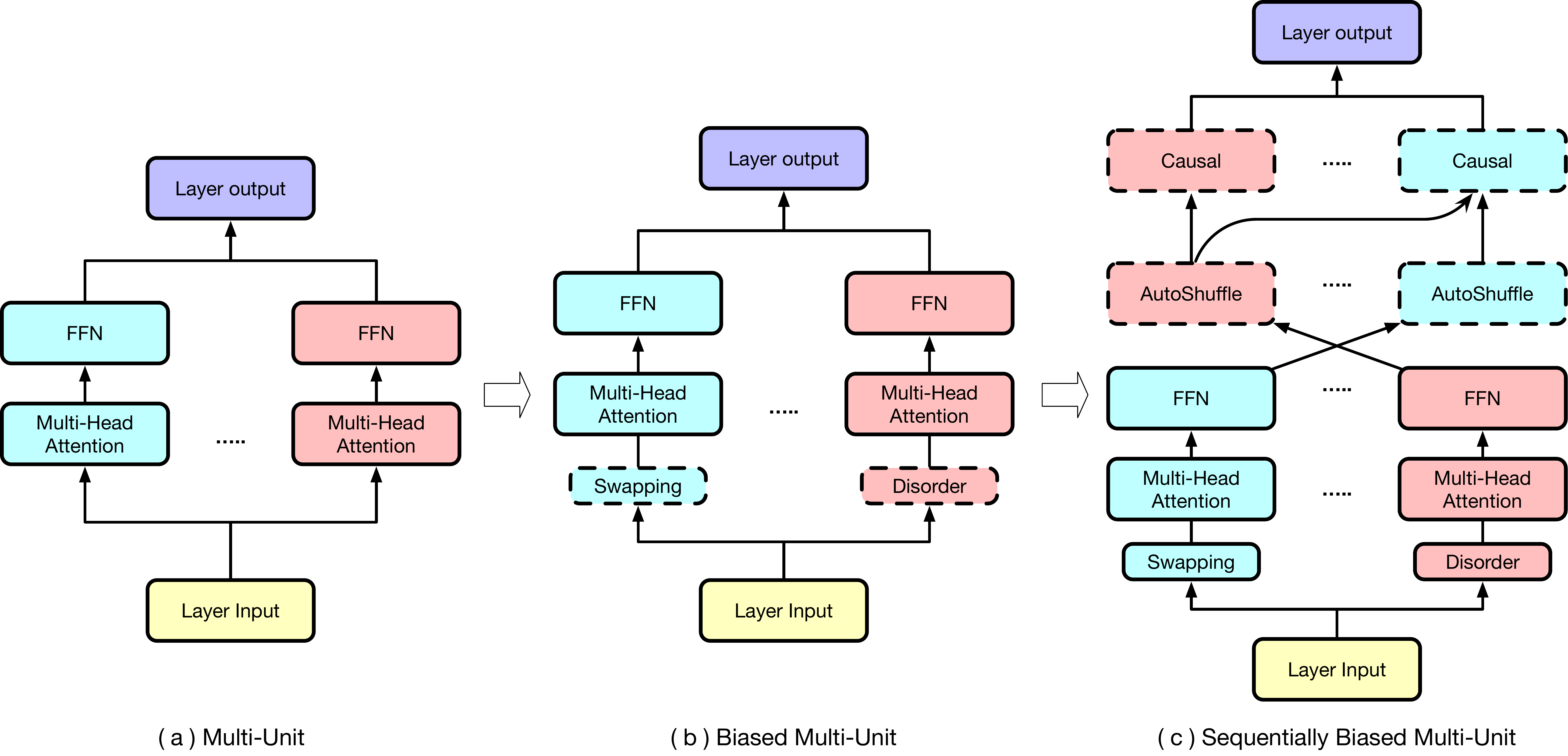}
    \caption{ Layer architecture for Multi-Unit Transformer. White arrow indicates model change from Multi-Unit to Biased Multi-Unit, to Sequentially Biased Multi-Unit, with dashed lines representing the newly added modules.}
    \label{fig:model-arch}
\vspace{-10pt}
\end{figure*}

\section{Model Layout}
\subsection{MUTE}
In this section, we briefly describe the Multi-Unit Transformer (MUTE), shown in Figure \ref{fig:model-arch}(a). 
As mentioned before, we take the bundle of cascading sub-components (i.e., attention modules and FFN) as a single unit.
By combining units in parallel, we have a basic Multi-Unit Transformer (MUTE) layer. 
Then, we follow the standard usage by stacking several MUTE layers to constitute our encoder and decoder.

In general, the encoder and decoder of the Transformer network share a similar architecture and can be improved with the same techniques. Without losing generality, we take the encoder as an example to further illustrate the MUTE. 
Given input $X^k$ of $k$-th layer, we feed it into $I$ identical units $\{F_1, \cdots, F_i, \cdots, F_I\}$ with different learnable parameters.
\begin{gather}
    s_i^k = \textup{SelfAttn}_i (X^k) + X^k, \\
    F_i^e(X^k) = s_i^k + \textup{FFN}_i(s_i^k),
\end{gather}
where $i$ denotes the $i$-th unit. After collecting outputs for all $I$ units, we combine them by a weighted sum,
\begin{gather}
    F^e(X^k) = \sum_{i \in (1, I)} \alpha_i \cdot F_i^e(X^k),
\end{gather}
where $\alpha_i \in \mathbb{R}^1$ represents the learnable weight for the $i$-th unit (Section \ref{sec:vis}) and $F^e(X^k) \in \mathbb{R}^{n \times d} $ is the final output for the $k$-th layer. 

\subsection{Biased MUTE}

The multi-unit setting for Transformer resembles the well-known ensemble techniques in machine learning fields, in that it also combines several different modules into one and aims for better performance. 
In that perspective, borrowed from the idea of bagging \cite{breiman1996bagging}, we propose to use biased units instead of identical units, which results in creating information gaps among units and makes them learn from each other.

More specifically, in training, we introduce a \emph{Bias-Module} to create biased units, as shown in Figure \ref{fig:model-arch}(b). For each layer, instead of giving the same inputs $X^k \in \mathbb{R}^{n \times d}$ to all units, we transform each input with corresponding type of noises (e.g., swap, reorder, mask), in order to force the model to focus on different parts of inputs:
\begin{gather}
    X_i^k = \textup{Bias}_i (X^k), \\
    F^e(X^k) = \sum_{i \in (1, I)} \alpha_i \cdot F_i^e(X_i^k),
\end{gather}
where $\textup{Bias}_i$ denotes the noise function for $i$-th unit. The noise operations \footnote{To avoid a distortion of input sequence, each operation is performed only once for corresponding bias module.} we investigated include,
\begin{compactitem}
    \item \textbf{Swapping}, randomly swap two input embeddings up to a certain range (i.e., 3). 
    \item \textbf{Disorder}, randomly permutate a subsequence within a certain length (i.e., 3).
    \item \textbf{Masking}, randomly replace one input embedding with a learnable mask embedding. 
\end{compactitem}
Note that, the identity mapping (i.e., no noise) can be seen as a special case of bias module and is included in our model design. 

Additionally, to get deterministic outputs, we disable the noises in the testing phase, which brings in the inconsistency between training and testing.
Hence, we propose a switch mechanism with a sample rate $p_\beta$ that determines whether to enable the bias module in training. This mechanism forces the model to adapt to golden inputs and mitigate the aforementioned inconsistency.

\subsection{Sequentially Biased MUTE}
Although the bias module guides units of learning from each other by formulating such information gaps, it still lacks explicit complementarity modeling, i.e., mitigating these gaps.
Here, based on the Biased MUTE, we propose a novel method to explicitly introducing a deep connection among units by utilizing the power of order (Figure \ref{fig:model-arch}(c)). 

\vspace{-5pt}

\paragraph{Sequential Dependency.}
Given the outputs from biased units $F^e_{i}(X_{i}^k)$, we permutate these outputs by a certain ordering function $p(i)$ (e.g., $i \in \{1,2,3,4\}$ to $ p(i) \in \{4,2,3,1\}$), 
\begin{gather}
    \{G^e_{i} = F^e_{p(i)}(X_{p(i)}^k) | i \in (1,I)\},
\end{gather}
where $G^e_{i}$ is the $i$-th permutated output.
The implementation of $p(i)$ will be illustrated later. 

Then, we explicitly model the complementariness among units by introducing sequential dependency. 
Specifically, we compute an accumulated sequence $\{\hat{G}^e_{i}| i \in (1, I)\}$ over the permutated outputs $G^e_{i}$,
\begin{gather}
    \hat{G_{i}^e} = \hat{G}_{i-1}^e + G_{i}^e,
\end{gather}
where $\hat{G}_{i}^e \in \mathbb{R}^{n \times d}$ is the $i$-th \emph{accumulated} output, and $\hat{G}_0^e = 0$.
Through this sequential dependency, each permutated output $G_{i}^e$ learns the residual of previous \emph{accumulated} outputs $\hat{G}_{i-1}^e$ \cite{he2016deep} and serves as a complement to previous accumulated outputs.

Finally, we normalize this accumulated sequence to keep the output norm stable and fuse all \emph{accumulated} outputs.
\begin{gather}
    F^e(X^k) = \sum_{i \in (1, I)} \alpha_{i} \cdot \frac{\hat{G}_{i}^e}{i}.
\end{gather}

\vspace{-5pt}

\paragraph{Autoshuffle.} 
Until now, we have modeled the sequential dependency between each of the units. 
The only problem left is how to gather a proper ordering of units. 
We propose to use the AutoShuffle Network \cite{lyu2019autoshufflenet}. 
Mathematically, shuffling with specific order equals to a multiplication by a permutation matrix (i.e., every row and column contains precisely a single 1 with 0s elsewhere). 
Nevertheless, a permutation matrix only contains discrete values and can not be optimized by gradient descent. Therefore, we use a continuous matrix $M \in \mathbb{R}^{I \times I}$ with non-negative values $M_{i,j} \in (0, 1), i, j \in (1, I)$ to approximate the discrete permutation matrix.

Particularly, $M$ is regarded as a learnable matrix and is used to multiply the outputs of units,
\begin{gather}
    [\cdots; F_{p(i)}^e(X^k_{p(i)}); \cdots] = M^{\top} \times [\cdots; F_{i}^e(X_{i}^k); \cdots],
\end{gather}
where $[\cdot;\cdot]$ means the concatenation operation.

To ensure $M$ remains an approximation for the permutation matrix during training, we normalize $M$ after each optimization step.
\begin{gather}
    M_{i,j} = \textup{max}(M_{i,j}, 0), \\
    M_{i,j} = \frac{M_{i,j}}{\sum_{\hat{i}} M_{\hat{i},j}}, M_{i,j} = \frac{M_{i,j}}{\sum_{\hat{j}} M_{i,\hat{j}}}.
\end{gather}

Then, we introduce a Lipschitz continuous non-convex penalty, as proposed in \citet{lyu2019autoshufflenet} to guarantee $M$ converge to a permutation matrix.
\begin{equation}
    \begin{aligned}
    \mathcal{L}_p =& \sum_{i=1}^I [\sum_{j=1}^I |M_{i,j}| - (\sum_{j=1}^I M_{i,j}^2 )^{\frac{1}{2}} ]  \\
        &+ \sum_{j=1}^I [\sum_{i=1}^I |M_{i,j}| - (\sum_{i=1}^I M_{i,j}^2 )^{\frac{1}{2}} ],
    \end{aligned}
\end{equation}
Please refer to Appendix \ref{sec:penalty_proof} for proof and other details.

Finally, the penalty is added to cross-entropy loss defined in equantion (\ref{eq:loss}) as our final objective, 
\begin{gather}
    \mathcal{L} = \mathcal{L}_{CE} + \lambda \sum_k \mathcal{L}^k_p,
\end{gather}
where $\lambda$ is a hyperparameter to balance two objectives and $\mathcal{L}^k_p$ is the penalty for the $k$-th layer.

\section{Experimental Settings}
In this section, we elaborate our experimental setup on three widely-studied Neural Machine Translation tasks, NIST Chinese-English (Zh-En), WMT'14 English-German (En-De) and WMT'18 Chinese-English. 

\begin{table*}
    \centering
    \renewcommand{\arraystretch}{1.1}
    \setlength{\tabcolsep}{1.5mm}{
    \scalebox{0.90}{
    \begin{tabular}{ l | c | l | l | l | l |l|c }
     \toprule
    \textbf{System} & \textbf{\#Param.} & \textbf{\emph{Valid}}& \textbf{\emph{MT02}}& \textbf{\emph{MT03}}& \textbf{\emph{MT04}}& \textbf{\emph{MT08}}& $\Delta$ \\ 
    \hline

    \multicolumn{8}{c}{\em Existing NMT systems} \\
    \hline
    \citet{wang2018switchout} (Base) & 95M & 45.47 & 46.31  & 45.30 & 46.45 & 35.66 & $-$\\
    \citet{cheng2018towards} (Base) & 95M & 45.78  & 45.96 & 45.51 & 46.49 &  36.08 & $-$\\
    \citet{cheng2019robust} (Base) & 95M & 46.95 &47.06 &46.48&47.39& 37.38 & $-$ \\
    \hline
    \multicolumn{8}{c}{\em Baseline NMT systems} \\
    \hline
    Transformer (Base) & 95M & 45.79 & 46.34 & 45.32 &47.92&36.16& \emph{ref}  \\
    Transformer + Relative (Base) & 95M & 46.57 &46.64&45.67&47.26& 38.03& +0.53 \\
    Transformer + Relative (Big) & 277M &46.52 & 47.23 &46.43&48.35& 37.31& +0.86 \\

    \hline
    \multicolumn{8}{c}{\em Our NMT systems} \\
    \hline
    MUTE (Base, 4 Units) & 152M &47.23$^{\dagger}$ &47.38$^{\dagger}$&46.24$^{\dagger}$&47.81$^{\dagger}$&38.48& +1.12 \\
    \quad + Bias (Base, 4 Units)  & 152M & 47.45$^{\dagger\dagger}$ & 47.40$^{\dagger\dagger}$ & \textbf{47.11}$^{\dagger\dagger}$ & \textbf{48.44}$^{\dagger\dagger}$ &38.31 & +1.44 \\
    \quad + Bias + Seq. (Base, 4 Units)  & 152M & \textbf{47.80}$^{\dagger\dagger}$ & \textbf{47.72}$^{\dagger\dagger}$ & 46.60$^{\dagger\dagger}$ & 48.30$^{\dagger\dagger}$ & \textbf{38.70} & \textbf{+1.52} \\
    
    \hline
    
    \hline
    \end{tabular}}}
    \caption{\label{tab:zhen} Case-insensitive BLEU scores (\%) of NIST Chinses-English (Zh-En) task. For all models with \textbf{MUTE}, we use four units. 
    \textbf{\#Params.} means the number of learnable parameters in the model.
    $\Delta$ denotes the average BLEU improvement over dev set and test sets, compared with the ``Transformer (Base)''. \textbf{Bold} represents the best performance.  
    ``$\dagger$'': significantly better than ``Transformer + Relative (Base)'' ($p < 0.05$); ``$\dagger\dagger$'': significantly better than ``Transformer + Relative (Base)'' ($p < 0.01$).}
    \vspace{-10pt}
\end{table*}

\vspace{-5pt}
\paragraph{Datasets.}
For NIST Zh-En task, we use 1.25M sentences extracted from LDC corpora\footnote{The corpora include LDC2002E18, LDC2003E07, LDC2003E14, Hansards portion of LDC2004T07, LDC2004T08 and
LDC2005T06.}. 
To validate the performance of our model, we use the NIST 2006 (MT06) test set with 1664 sentences as our validation set.
Then, the NIST 2002 (MT02), 2003 (MT03), 2004 (MT04), 2008 (MT08) test sets are used as our test sets, which contain 878, 919, 1788 and 1357 sentences, respectively. 

For the WMT'14 En-De task, following the same setting in \citet{vaswani2017attention}, we use 4.5M preprocessed data, which has been tokenized and split using byte pair encoded (BPE) \cite{sennrich-etal-2016-neural} with 32k merge operations and a shared vocabulary for English and German. 
We use \emph{newstest2013} as our validation set and \emph{newstest2014} as our test set, which contain 3000 and 3003 sentences, respectively. 

For the WMT'18 Zh-En task, we use 18.4M preprocessed data, which is also tokenized and split using byte pair encoded (BPE) \cite{sennrich-etal-2016-neural}. We use \emph{newstest2017} as our validation set and \emph{newstest2018} as our test set, which contains 2001 and 3981 sentences, respectively.

\vspace{-5pt}
\paragraph{Evaluation.}
For evaluation, we train all the models with maximum 150k/300k/300k steps for NIST Zh-En, WMT En-De and WMT Zh-En, respectively, and we select the model which performs the best on the validation set and report its performance on the test sets. 
We measure the case-insensitive/case-sensitive BLEU scores using \emph{multi-bleu.perl} \footnote{\url{https://github.com/moses-smt/mosesdecoder/blob/master/scripts/generic/multi-bleu.perl}} with the statistical significance test \cite{koehn-2004-statistical} \footnote{\url{https://github.com/moses-smt/mosesdecoder/blob/master/scripts/analysis/bootstrap-hypothesis-difference-significance.pl}} for NIST Zh-En and WMT'14 En-De, respectively.
For WMT'18 Zh-En, we use case sensitive BLEU scores calculated by Moses \emph{mteval-v13a.pl} script \footnote{\url{https://github.com/moses-smt/mosesdecoder/blob/master/scripts/generic/mteval-v13a.pl}} . 

\vspace{-5pt}
\paragraph{Model and Hyper-parameters.}
For all our experiments, we basically follow two model settings illustrated in \citep{vaswani2017attention}, namely \emph{Transformer-Base} and \emph{Transformer-Big}.
In \emph{Transformer-Base}, we use 512 as hidden size, 2048 as filter size and 8 heads in multihead attention.
In \emph{Transformer-Big}, we use 1024 as hidden size, 4096 as filter size, and 16 heads in multihead attention.

Besides, since noise types like swapping and reordering are of no effect on Transformer models with absolute position information, the MUTE models are implemented using relative position information \cite{shaw2018self}. 
In addition, we only apply multi-unit methods to encoders, provided that the encoder is more crucial to model performance \cite{wang2019learning}. 
All experiments on MUTE models are conducted with \emph{Transformer-Base} setting.
For the basic MUTE model, we use four identity units. As for the Biased MUTE and Sequentially Biased MUTE, we use four units including one identity unit, one swapping unit, one disorder unit and one masking unit.  
The sample rate $p_\beta$ is set to 0.85.
For more implementation details and experiments on sample rate, please refer to Appendix \ref{sec:impl} and \ref{sec:sample}.

\section{Results}
Through experiments, we first evaluate our model performance (Section \ref{sec:zhen} and \ref{sec:ende}). 
Then, we analyze how each part of our model works (Section \ref{sec:abla} and \ref{sec:div}). 
Finally, we conduct experiments to further understand the behavior of our models (Section \ref{sec:num} and \ref{sec:vis}).

\subsection{Results on NIST Chinses-English}
\label{sec:zhen}
As shown in Table \ref{tab:zhen}, we list the performance of our re-implemented Transformer baselines and our approaches. We also list several existing strong NMT systems reported in previous work to validate the effectiveness of our models. 
By investigating results in Table \ref{tab:zhen}, we have the following observations. 

First,
compared with existing NMT systems, our re-implemented Transformers are strong baselines. 

Second, all of our approaches substantially outperform our baselines, with improvement ranging from 1.12 to 1.52 BLEU points. Comparing our methods to ``Transformer + Relative (Base)'', our best approach (i.e., ``MUTE + Bias + Seq. (Base, 4 Units)'') still achieves a significant improvement of 1.0 BLEU points on multiple test sets.

\begin{table}[!tbp]
    \centering
    \renewcommand{\arraystretch}{1.1}
    \setlength{\tabcolsep}{0.5mm}{
    \scalebox{0.90}{
    \begin{tabular}{ l | r | l | c }
     \toprule
    \textbf{System} & \textbf{\#Param.} &\textbf{BLEU}& $\Delta$ \\ 
   \hline
    \multicolumn{4}{c}{\em Existing NMT systems} \\
    \hline
    \citet{vaswani2017attention} (Base) & 81M  & 27.3 & ref. \\
    \citet{bapna2018training} (Base, 16L) & 137M & 28.0 & - \\
    \citet{wang2019learning} (Base, 30L) & 137M & 29.3 & - \\
    \hline 
    \citet{vaswani2017attention} (Big) & 251M  & 28.4 & - \\
    \citet{chen2018best} (Big)  & 379M & 28.5 & - \\
    \citet{he2018layer} (Big)  & *251M &  29.0 & - \\
    \citet{shaw2018self} (Big) & *251M &  29.2 & - \\
    \citet{ott2018scaling} (Big) & 251M &  29.3 & - \\
    \hline
    \multicolumn{4}{c}{\em Existing Multi-Unit Style NMT Systems} \\
    \hline
    \citet{ahmed2017weighted} (Base) & *81M &  28.4 & - \\ 
    \citet{li2019neuron} (Base) & 118M & 28.5& - \\
    \citet{li2018multi} (Base) & *81M & 28.5& - \\
    \hline
    \multicolumn{4}{c}{\em Baseline NMT systems} \\
    \hline
    Transformer (Base) & 81M &  27.4 & ref. \\
    Transformer + Relative (Base) & 81M &  28.2 & +0.8 \\
    Transformer + Relative (Big) & 251M &  28.6 & +1.2 \\
    \hline
    \multicolumn{4}{c}{\em Our NMT systems} \\
    \hline
    MUTE (Base, 4 Units) & 130M &  28.8$^{\dagger\dagger}$ & +1.4 \\
    \quad + Bias (Base, 4 Units) & 130M &  29.1$^{\dagger\dagger}$ & +1.7 \\
    \quad + Bias + Seq. (Base, 4 Units) & 130M  & \textbf{29.3}$^{\dagger\dagger}$& \textbf{+1.9} \\
    \hline
    \end{tabular}}}
    \caption{\label{tab:ende} Case-sensitive BLEU scores (\%) of WMT'14 English-German (En-De) task. 
    $\Delta$ denotes the improvement over \emph{newstest2014}, compared with Transformer (Base). * denotes an estimated value. ``$\dagger\dagger$'': significantly better than ``Transformer + Relative (Base)'' ($p < 0.01$).} 
    \vspace{-8pt}
  \end{table}

Third, among our approaches, we find that, even though our basic MUTE model has already surpassed existing strong NMT systems and our baselines, the bias module and sequential dependency can further boost the performance (i.e., from +1.12 to +1.52), which demonstrates that introducing complementariness does help the Multi-Unit Transformers.

Fourth, we find it interesting that compared with the ``Transformer + Relative (Big)'', the basic ``MUTE (Base, 4 Units)'' can achieve better BLEU performance with only 54\% of parameters, which indicates that our multi-unit approaches can leverage parameters more effectively and efficiently.

\subsection{Results on WMT'14 English-German}
\label{sec:ende}

The results on WMT'14 En-De are shown in Table \ref{tab:ende}.
We list several competitive NMT systems for comparison, which are divided into models based on \emph{Transformer-Base} and models based on \emph{Transformer-Big}. 

First of all, our models show significant BLEU improvements over two baselines in the \emph{Transformer-Base} setting, ranging from +1.4 to +1.9 for ``Transformer (Base)'' and from +0.6 to +1.1 for ``Transformer+Relative (Base)''.
That proves our methods perform consistently across languages and are still useful in large scale datasets.

Next, among our own NMT methods, the sequentially biased model further improves the BLEU performance over our strong Multi-Unit model (from 28.8 to 29.3),
which is consistent with our findings in the Zh-En\footnote{We find that improvements on the larger scale WMT'14 En-De dataset is bigger than that on NIST Zh-En. We attribute this phenomenon to the overfitting problem caused by the small Zh-En dataset.}
task and further proves the power of complementariness.

Finally, compared with the existing NMT systems, we find that our models achieve comparable / better performance with much fewer parameters. 
The only exception is \citep{wang2019learning}, which learns a very deep (30 layers) Transformer. 
We regard these deep Transformer methods as orthogonal methods to ours, and it can be integrated with our MUTE models in future work. 
Additionally, we list several systems related to our multi-unit setting (``\emph{Existing Multi-Unit Style NMT Systems}''), with diversity modeling or features space composition.
As shown, MUTE models also outperform these methods, demonstrating the superiority of our methods in diverse and complementary modeling.

\begin{table}[!tbp]
    \centering
    \renewcommand{\arraystretch}{1.1}
    \setlength{\tabcolsep}{0.5mm}{
    \scalebox{0.90}{
    \begin{tabular}{ l | r | c | c }
     \toprule
    \textbf{System} & \textbf{\#Param.} &\textbf{BLEU}& $\Delta$ \\ 
    \hline
    Transformer (Base) & 81M &  23.4 & ref. \\
    Transformer + Relative (Base) & 81M &  23.8 & +0.4 \\
    \hline
    MUTE + Bias + Seq. & 130M  & \textbf{24.5}& \textbf{+1.1} \\
    \hline
    \end{tabular}}}
    \caption{\label{tab:wmt-zhen} Case-sensitive BLEU scores (\%) of WMT'18 Chinese-English (Zh-En) task. }
    \vspace{-8pt}
  \end{table}

\begin{table}[!t]
    \centering
    \setlength{\tabcolsep}{2.0mm}{
    \scalebox{0.95}{
    \begin{tabular}{l | c }  
    \toprule
    \textbf{Methods}  & \textbf{BLEU Score}  \\
    \midrule
    MUTE + Bias + Seq.  & \textbf{47.80}\\
    \midrule
    \,- identity unit. &  47.01 (-0.79)\\
    \,- swapping unit. &  47.13 (-0.67) \\
    \,- disorder unit. &  47.15 (-0.65)\\
    \,- mask unit. &  47.36 (-0.44) \\
    \hline 
    \,- bias module & 47.28 (-0.52)  \\
    \,- sequential dependency & 47.45 (-0.35)  \\
    \bottomrule
    \end{tabular}}}
    \caption{Ablation study for BLEU scores (\%) over the NIST Zh-En validation set. }
    \label{tab:ablation}
    \vspace{-10pt}
\end{table}

\subsection{Results on WMT'18 Chinese-English}
\label{sec:wmt-zhen}

In this section, we represent our results on WMT18 Chinese-English. The results are shown in Table \ref{tab:wmt-zhen}. As we can see, our MUTE model still strongly outperforms the baseline ``Transformer (Base)'' and  ``Transformer+Relative (Base)'' with +1.1 and +0.7 BLEU points.
Noting that WMT'18 Zh-En has a much larger dataset (18.4M), and these findings proves that our model perform consistently well with different size of datasets. 


\subsection{Ablation Study}
\label{sec:abla}
In this section, we conduct the ablation study to verify each part of our proposed model. 
The results are shown in Table \ref{tab:ablation}. 
From our strongest Sequentially Biased model, we remove each of the four different units to validate which unit contributes the most to the performance. 
Then, we remove the bias module and sequential dependency independently to investigate each module's behavior.

We come to the following conclusions:

(1) All units make substantial contributions to ``MUTE + Bias + Seq.'', ranging from 0.44 to 0.79, proving the effectiveness of our design.

(2) Among all units, the identity unit contributes most to our performance, which is consistent with our intuition that the identity unit should be responsible most for complementing other biased units.

(3) The bias module and sequential dependency both contribute much to our Multi-Unit Transformers. We find it intriguing that, without the bias module, sequential dependency only provides marginal improvements. We conjecture that the complementary effect becomes minimal with no information gap among units.

\subsection{Comparison with Averaging Checkpoints}
\label{sec:avg_ckpt}
As we mentioned before, our MUTE models are inspired by ensembling methods. Therefore, it is necessary to compare our model with representative ensemble methods, e.g., averaging model checkpoints. Here, the comparison results of our MUTE model and averaging checkpoints in NIST Zh-En are shown in Table \ref{tab:avg}.
Since averaging checkpoints often leads to better generalization, we report the average BLEU scores over all test sets.

We adopt two settings, namely averaging the last several (i.e., 5) checkpoints and averaging over models initialized with different seeds. The experiment with different initialization seeds fails. We conjecture the reason is that different seeds make models fall in different sub-optimals, and brutally combining them together makes the model perform badly. 
Then we average checkpoints over the last 5 saves, which gives us 44.97 BLEU points, which only outperforms the best checkpoint marginally (+0.14 in average), and MUTE performs much better (45.43 and 45.82 BLEU points). 
Specificially, our naive MUTE model suprasses the averaginig checkpoint method, and the sequential ordering and bias module enable a better interaction over different units.

\begin{table}[!t]
    \centering
    \setlength{\tabcolsep}{2.0mm}{
    \scalebox{0.95}{
    \begin{tabular}{l | c } 
    \toprule
    \textbf{Methods}  & \textbf{BLEU Score}  \\
    \midrule
    Transformer + Relative  & 44.83 \\
    \quad + average last 5 saves & 44.97 \\
    \quad + average 5 seeds & \emph{Fails} \\
    \midrule
    MUTE & 45.43 \\
    MUTE + Bias + Seq.  & \textbf{45.82}\\
    \bottomrule
    \end{tabular}}}
    \caption{Comparison with averaging checkpoints in terms of BLEU scores (\%) over the NIST Zh-En datasets. The reported BLEU scores is the average of all test sets.}
    \label{tab:avg}
    \vspace{-10pt}
\end{table}


\subsection{Quantitative Analysis of Model Diversity}
\label{sec:div}

Here, we empirically investigate which granularity should be used for better diversity among units.
To verify the impact of multiple units compared with the single unit, 
we evaluate three different models:
\begin{compactitem}
    \item \textbf{4 Self. + 4 FFN}, the model with four different self-attention modules and four different FFNs.
    \item \textbf{4 Self. + 1 FFN}, the model with four self-attention modules and one shared FFN.
    \item \textbf{1 Self. + 4 FFN}, the model with one shared self-attention module and four different FFNs.
\end{compactitem}
To control variables, these models include neither bias module nor sequential dependency.

For each model, we evaluate the diversity among units for three outputs: (1) the outputs of self-attention modules, (2) the attention weights of self-attention modules, (3) the outputs of FFN modules. 
The diversity scores are computed by the exponential of the negative cosine distance among units, the same as proposed in \cite{li2018multi}. 
\begin{gather}
    \textup{DIV} = \textup{exp}(- \frac{o_i^{\top} o_j}{|o_i| \cdot |o_j|}),
\end{gather}
where $\textup{DIV}  \in (0, 1)$ represents the diversity score for module outputs $o_i \in \mathbb{R}^d$ and $o_j \in \mathbb{R}^d$.
The results are shown in Table \ref{tab:div}.

\begin{table}
    \centering
    \setlength{\tabcolsep}{0.5mm}{
    \scalebox{0.90}{
    \begin{tabular}{c|c|c|c}  
    \toprule
    \textbf{Methods}  & \textbf{Att Pos.} &\textbf{Att Sub.}&\textbf{FFN Sub.}\\
    \midrule
    MUTE + Bias + Seq.  & \textbf{0.475}  & \textbf{0.602} &\textbf{0.460}\\
    \midrule
    4 Self. + 4 FFN    & 0.447  & 0.576  & 0.449\\
    4 Self. + 1 FFN    & 0.420   & 0.520  & 0.404 \\
    1 Self. + 4 FFN    & -  & - & 0.381\\
    \bottomrule
    \end{tabular}}}
    \caption{Quantitative analysis on diversity scores.  \textbf{Att Pos.}, \textbf{Att Sub.} and \textbf{FFN Sub.} denote the diversity scores for self-attention weights, self-attention outputs and FFN outputs, respectively.
    A larger score means better diversity.}
    \vspace{-10pt}
    \label{tab:div}
\end{table}

Above all, we find that multiple FFN layers can introduce diversity. As seen, ``1 Self. + 4 FFN'' produces a 0.381 diversity score on ``FFN Sub.''. Since we use a shared self-attention layer, which brings no diversity in the input-side of FFN layers, the difference is only brought by different FFN modules. 
We think the reason is that the RELU activation inside the FFN module serves as selective attention to filter out input information.

Next, ``4 Self. + 1 FFN'' achieves 0.420 and 0.520 diversity scores for self-attention modules, which indicates that multiple self-attention modules focus on different parts of the input sequence and lead to diverse outputs. 

Then, ``4 Self. + 4 FFN'' has higher diversity scores than ``4 Self. + 1 FFN'' and ``1 Self. + 4 FFN'', which verifies our choice of using a combination of self-attention module and FFN as a basic unit.

Finally, concerning the diversity scores among all models, we find that our full model with biased inputs and sequential dependency achieves the best diversity scores for all three outputs, which also achieves the best BLEU scores in previous experiments. 

\subsection{Effects on the Number of Units}
\label{sec:num}
Another concern is how the MUTE models perform when increasing the number of units.
Thus, in this section, we empirically investigate the effects on the number of units, in terms of model performance and inference speed. 
Here we use the basic Multi-Unit Transformer\footnote{The number of combinations of different noise types is exponentially large. Moreover, introducing sequential dependency and bias module hardly affects the inference speed.}.


\begin{figure}
    \centering     
    \subfigure[Effects on BLEU]{
        \label{fig:bleu_change}
        \begin{minipage}{.48\columnwidth}
        \includegraphics[width=\columnwidth]{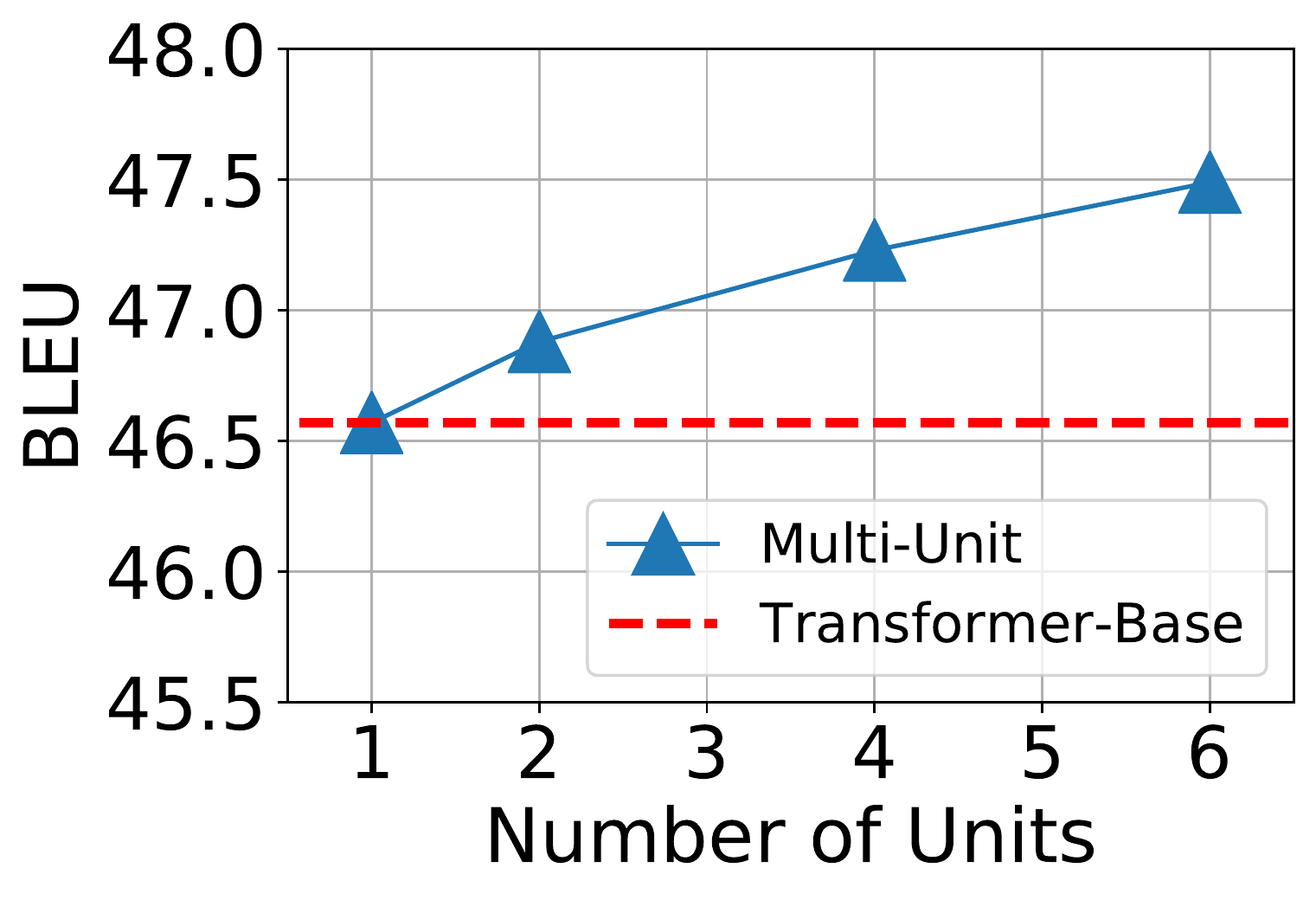}
        \end{minipage}
    }%
    \subfigure[Effects on Speed]{
        \label{fig:speed_change}
        \begin{minipage}{.48\columnwidth}
        \includegraphics[width=\columnwidth]
        {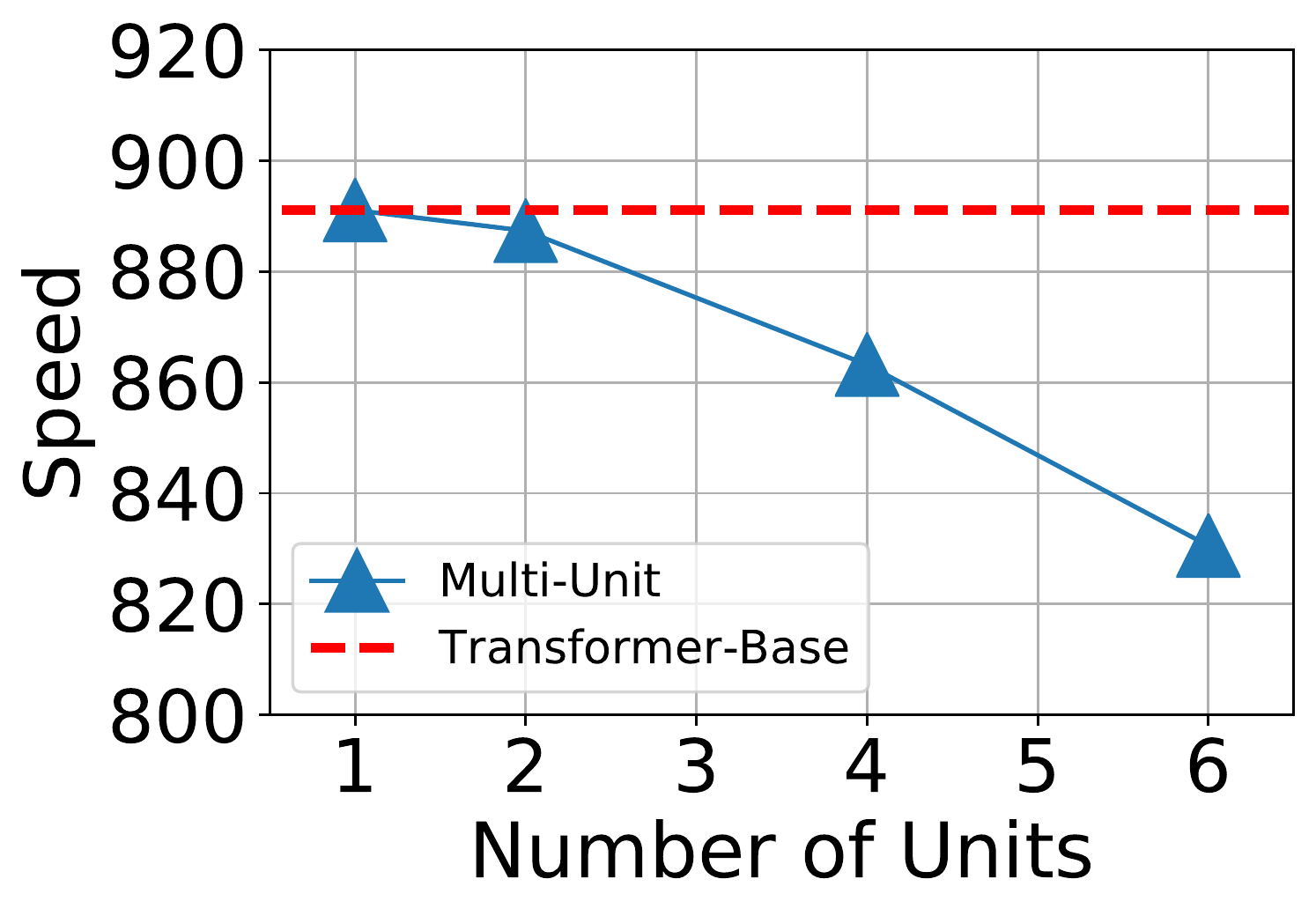}
        \end{minipage}
        }
    \vspace{-10pt}
    \caption{The effect on the number of units: (a) BLEU scores (\%) against number of units used in the MUTE model. (b) Inference speed (tokens per second) on GPU against number of units used in the MUTE model. Experiments are conducted on the valid set of NIST Zh-En with decoding batch size = 30 and beam size = 4.}
    \vspace{-10pt}
\end{figure}

As shown in Figure \ref{fig:bleu_change} and \ref{fig:speed_change}, increasing the number of units from 1 to 6 yields consistent BLEU improvement (from 46.5 to 47.5) with only mild inference speed decrease (from 890 tokens/sec to 830 tokens/sec). 
Besides, our model with four unit used in other experiments is faster than Transfomrer-Big (863 tokens/sec vs 838 tokens/sec).
These results prove the computational efficiency of our MUTE model. 
We attribute this mild speed decrease (about 3.1\% for four units and 
6.7\% for six units) for two reasons. 
First, the multi-unit model is naturally easy for parallelization. 
Each unit can be computed independently without waiting for other functions to finish. 
Second, we only widen the encoder, which is only computed once for each sentence translation.

\subsection{Visualization}
\label{sec:vis}
\begin{figure}
    \centering     
    \scalebox{0.95}{
    \subfigure[MUTE]{
        \label{fig:mute}
        \begin{minipage}{.48\columnwidth}
        \includegraphics[width=\columnwidth]{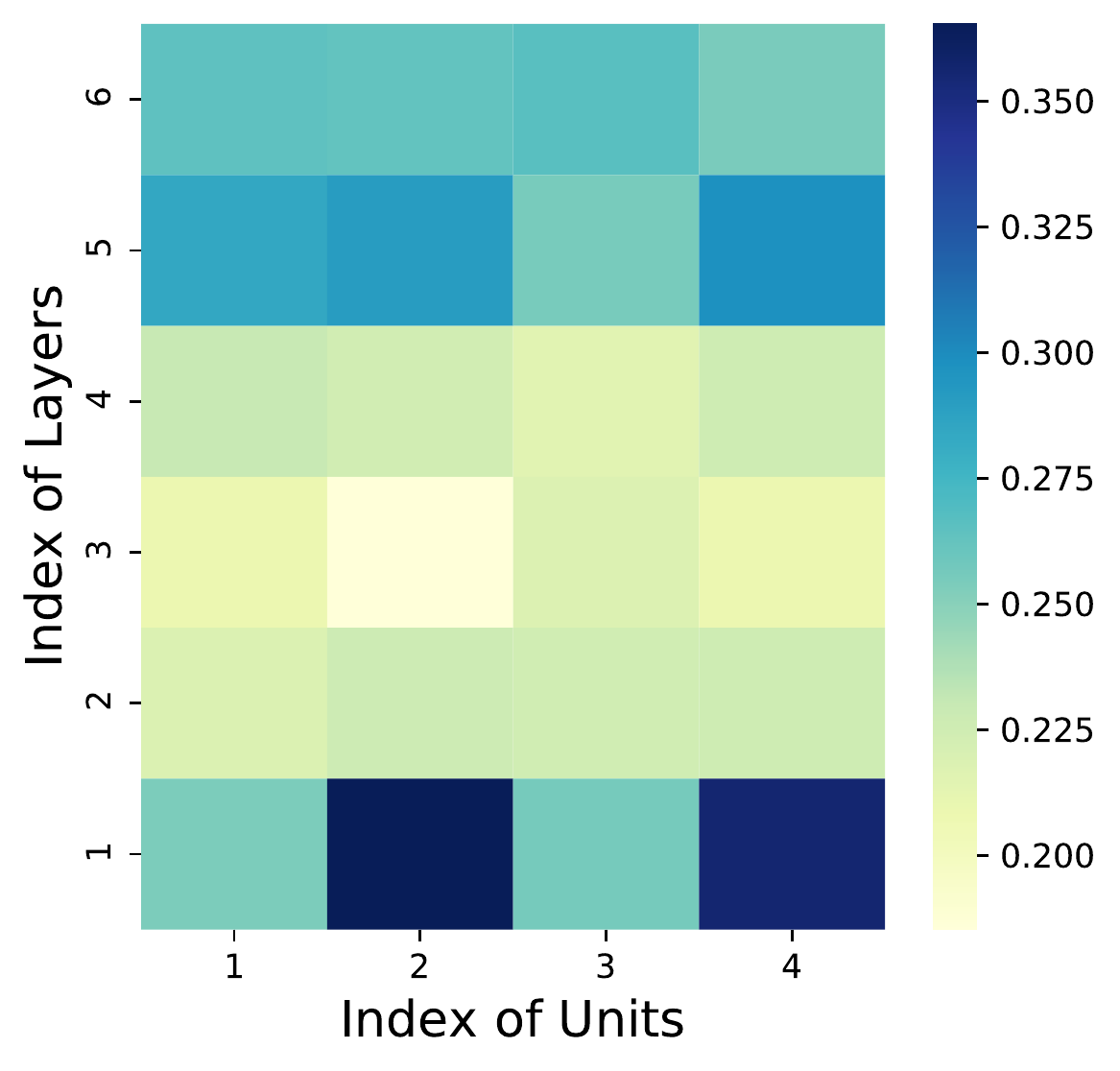}
        \end{minipage}
    }%
    \subfigure[MUTE + Bias]{ 
        \label{fig:bias_mute}
        \begin{minipage}{.48\columnwidth}
        \includegraphics[width=\columnwidth]
        {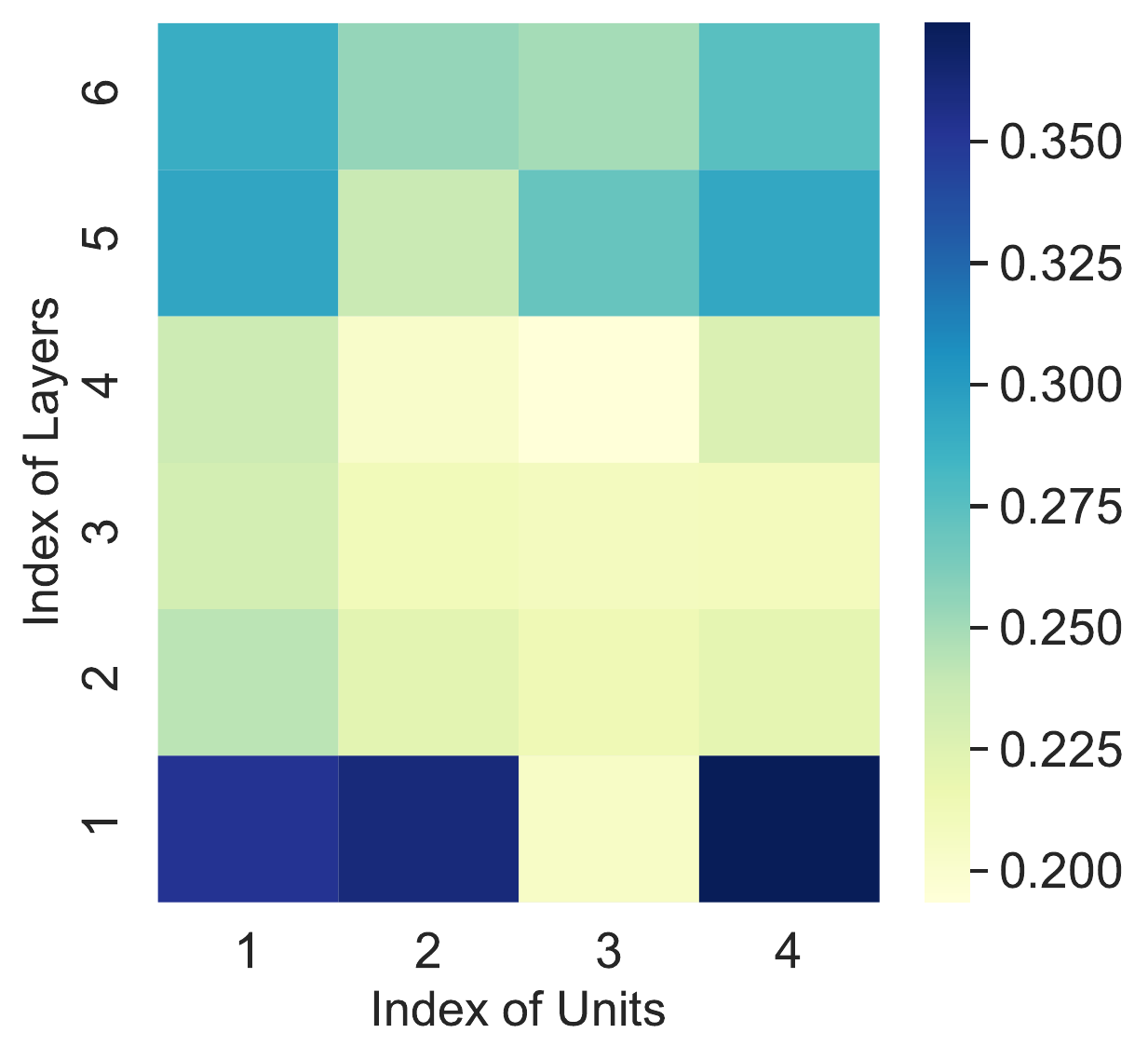}
        \end{minipage}
        }
    }
    \scalebox{0.95}{
    \subfigure[MUTE + Bias + Seq.]{
        \label{fig:seq_bias_mute}
        \begin{minipage}{.48\columnwidth}
        \includegraphics[width=\columnwidth]
        {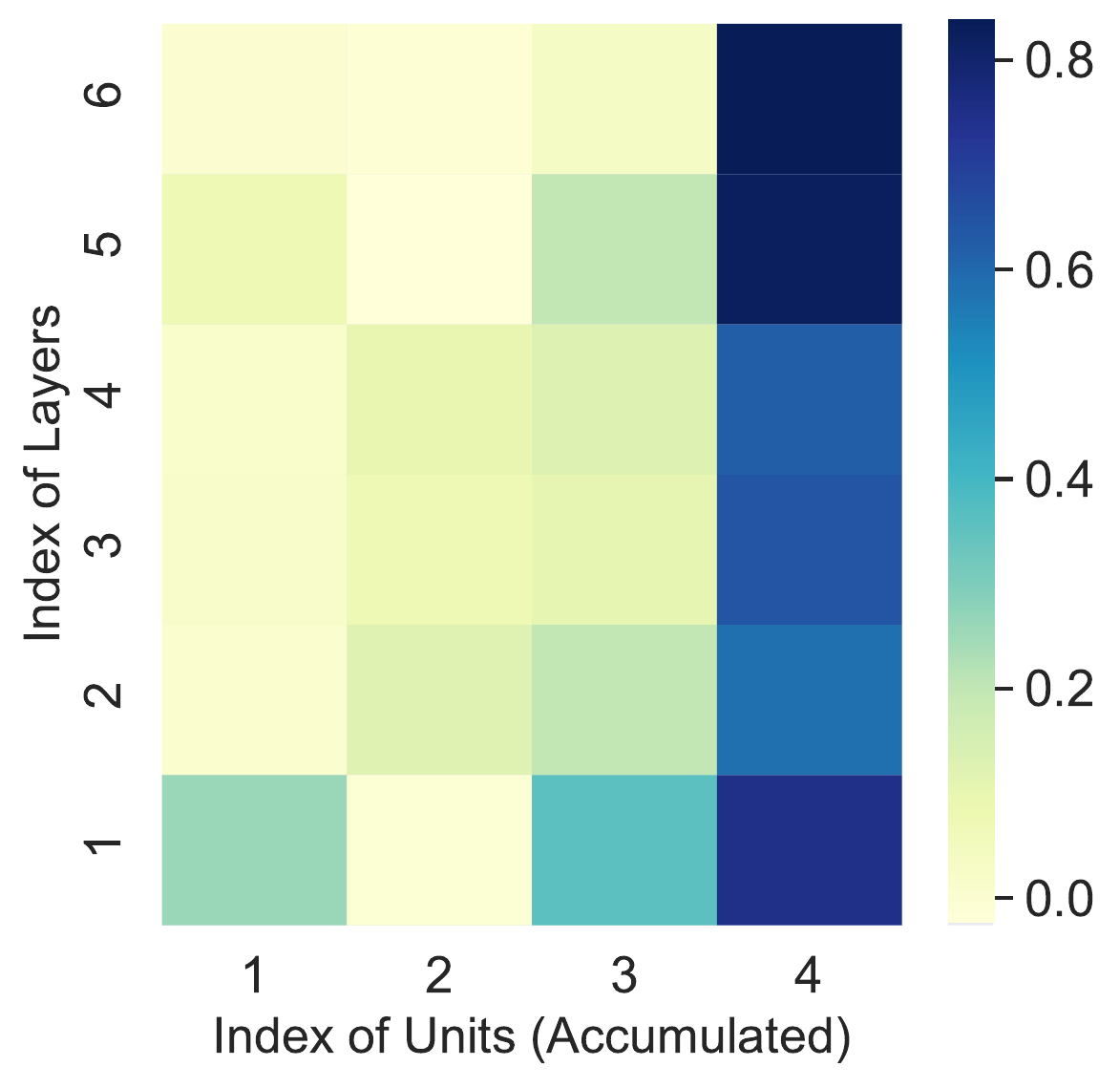}
        \end{minipage}
        }
    }
    \vspace{-0.3cm}
    \caption{Visualization of our models' learnable weights for units in each layer. }
    \vspace{-0.5cm}
    \label{fig:vis}
\end{figure}

We also present a visualization example of the learnable weights $\alpha$ for units, shown in Figure \ref{fig:vis}.

As we can see, learnable weights $\alpha$ show similar trends in ``MUTE'' and ``MUTE + Bias''. The weights for each unit within the same layer fall in a similar range (0.2 to 0.35), dispelling the worries that the biased units may be omitted or skipped. 
As for the ``MUTE + Bias + Seq.'', the weight distribution is very different. The weights nearly increase progressively when the unit index increases. Because the model weights are learned with back-propagation, the larger the weight, the more the model favors the corresponding unit. Thus, to some extent, this phenomenon demonstrates that the latter accumulated outputs are more potent than the preceding ones, and therefore, the latter single unit output complements the previous accumulation.

\section{Related Work}
Recently Transformer-based models \cite{vaswani2017attention,ott2018scaling,wang2019learning} become the de facto methods in Neural Machine Translation, owing to high parallelism and large model capacity.

Some researchers devise new modules to improve the Transformer model, including combining the transformer unit with convolution networks \cite{wu2019pay,zhao2019muse,lioutas2020time}, improving the self-attention architecture\cite{fonollosa2019joint,wang2019self,hao2019multi}, and deepening the Transformer architecture by dense connections\cite{wang2019learning}. 
Since our multi-unit framework makes no limitation about its unit, these models can be easily integrated into our multi-unit framework.

There are also some works utilizing the power of multiple modules to capture complex feature representations in NMT.
\citet{shazeer2017outrageously} use a vast network and a sparse gated function to select from multiple experts (i.e., MLPs).
\citet{ahmed2017weighted} train a weighted Transformer by replacing the multi-head attention by self-attention branches.
Nevertheless, these models ignore the modeling of relations among different modules.
Then, some multihead attention variants \cite{li2018multi,li2019neuron} introduce modeling of diversity or interaction among heads. However, complementariness is not taken into account in their approaches.
Our MUTE models differ from their methods in two aspects. First, we use a powerful unit with a strong performance in diversity (Section \ref{sec:div}). Second, we explicitly model the complementariness with bias module and sequential dependency. 

\section{Conclusion}
In this paper, we propose Multi-Unit Transformers for NMT to improve the expressiveness by introducing diverse and complementary units. 
In addition, we propose two novel techniques, namely bias module and sequential dependency to further improve the diversity and complementariness among units. 
Experimental results show that our methods can significantly outperform the baseline methods and achieve comparable / better performance compared with existing strong NMT systems. In the meantime, our methods use much fewer parameters and only introduce mild inference speed degradation, which proves the efficiency of our models.

\bibliography{anthology,emnlp2020}
\bibliographystyle{acl_natbib}


\appendix
\section{Implementation Details} \label{sec:impl}
All models are trained using \emph{Opennmt-py} framework \cite{opennmt}. 
The batch size for each GPU is set to 4096 tokens.
The beam size is set to 4, and the length penalty is 0.6 among all experiments. 
For the WMT'14 En-De and WMT'18 Zh-En task, all experiments are conducted using 4 NVIDIA Tesla V100 GPUs, while we use 2 GPUs for the NIST Zh-En task.
That gives us about 8k/16k/16k tokens per update for NIST Zh-En/WMT'14 En-De/WMT'18 Zh-En.
All models are optimized using Adam \cite{kingma2014adam} with $\beta_1 = 0.9$ and $\beta_2 = 0.998$, and learning rate is set to 1.0 for all experiments. 
Label smoothing is set to 0.1 for all three tasks. We use dropout of 0.4/0.2/0.1 for NIST Zh-en/WMT En-de/WMT Zh-en, respectively.
All our Transformer models contain 6 encoder layers and 6 decoder layers, following the standard setting in \cite{vaswani2017attention}.

All the hyperparameters are empirically set by our previous experiences or manually tuned. The criterion for selecting hyperparameters is the BLEU score on validation sets for both tasks. The average runtimes are one GPU day for NIST Zh-En and 4 GPU days for WMT'14 En-De and WMT'18 Zh-EN.

\section{Effect of Sample Rate}
\label{sec:sample}
\begin{figure}
    \centering
    \includegraphics[width=1.0\columnwidth]{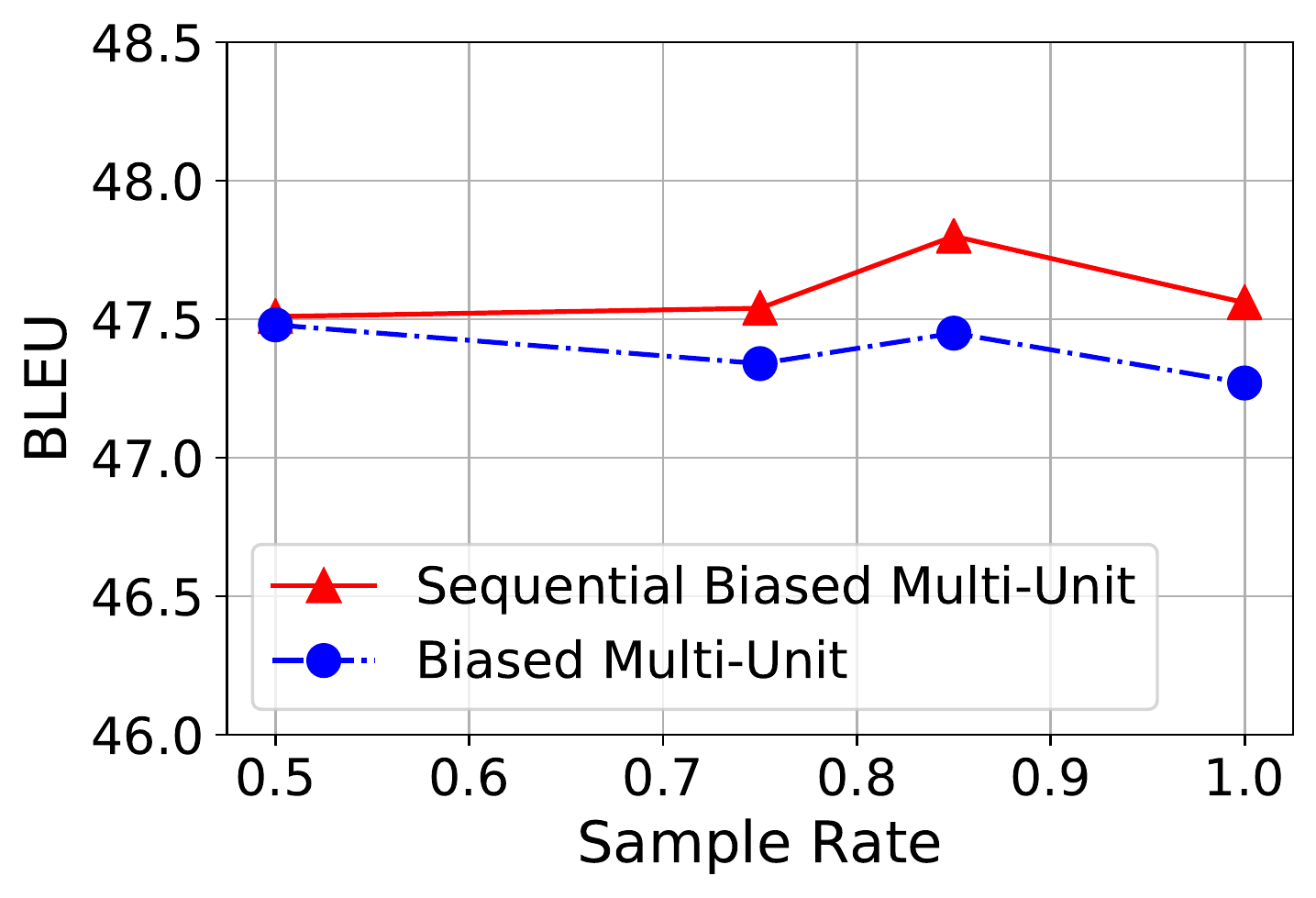}
    \caption{BLEU scores (\%) changes on NIST Zh-En valid set when increasing the sample rate $p_{\beta}$. Bigger value means a better chance to inject noises. 1.0 for sample rate means always injecting noises.}
    \label{fig:sample}
\end{figure}

As we mentioned above, we use sample rate to mitigate the incosistency between training and testing, and force the model to adapt to golden inputs. 
Here, we conduct hyper-parameter search for sample rate with 4 empirically set values, 0.5, 0.75, 0.85, 1.0. The results are shown in Figure \ref{fig:sample}. 

As seen, we make the following observations:
1. The trends of BLEU scores for sequentially biased model and biased model are very similar, when we increases the sample rate $p_\beta$. The best performance for both models appear when the sample rate is 0.85. Therefore, we set sample rate to 0.85 among all of our experiments.
2. Introducing sample rate is essential for both biased model and sequentially biased model. Compared with alway injecting noises, i.e., the sample rate $p_\beta = 1.0$, consistent improvements is observed for both models. BLEU scores increase from 47.27 to 47.48 (+0.21) for biased model, and from 47.56 to 47.80 (+0.24) for sequentially biased model.

The aforementioned observations prove that introducing sample rate when using bias module is an essential way to mitigate the inconsistency between training and testing.

\section{Case Study}
\begin{figure*}[!htbp]
    \centering
    \includegraphics[width=1.0\textwidth,trim=150 0 300 0,clip]{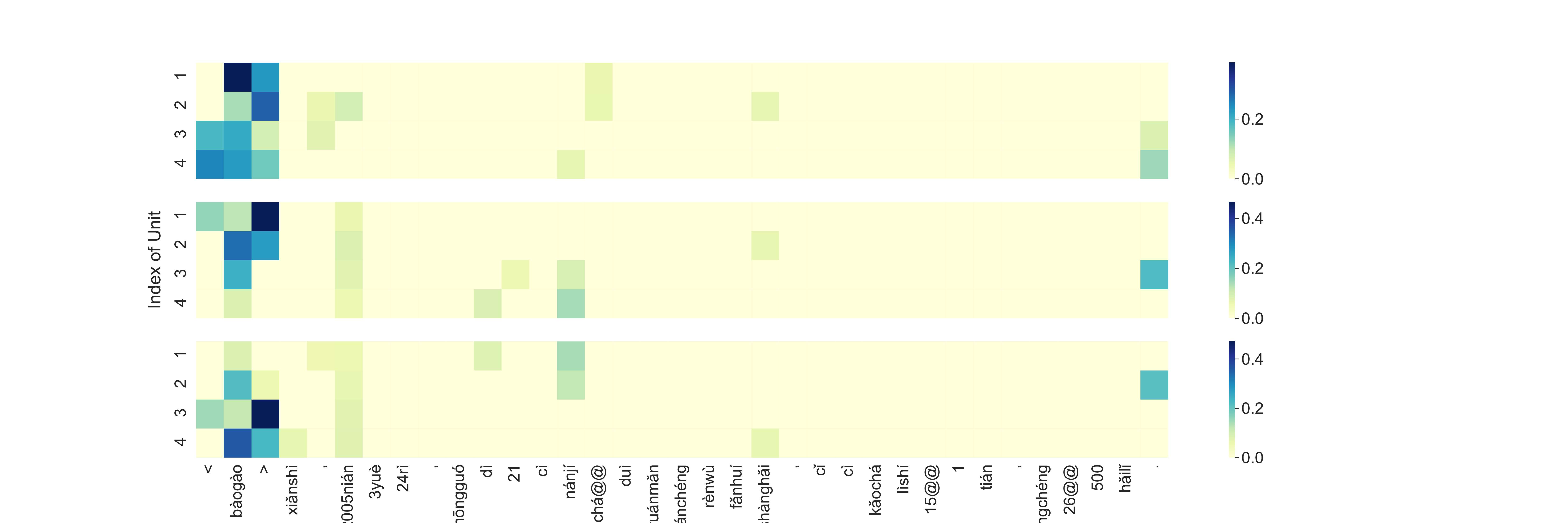}
    \vspace{-3pt}
    \caption{A visualization of example attention weights of the \emph{Multi-Unit}, \emph{Biased Multi-Unit} and \emph{Sequentially Biased Multi-Unit} models, top to bottom.
    We filter out the weights smaller than 0.05 for ease of understanding.}
    \label{fig:case}
    \vspace{-3pt}
\end{figure*}

We also provide a visualization example to show how our proposed methods improve complementariness among units.
As shown in Figure \ref{fig:case}, the top one is the attention weights of Basic Multi-Unit model. 
The attention weights for each of the units mainly focus on same parts of the inputs. 
In contrast to this phenomenon, the biased model and sequentially biased model attend with more diverse weights.
Moreover, we find that the sequentially biased model pays more attention to what the previous unit miss, which proves that our proposed method can actually encourage complementariness.

\section{Simple Proof for Lipschitz Penalty}
\label{sec:penalty_proof}
Here we give a intuitive proof why the Lipschitz penalty \cite{lyu2019autoshufflenet} would lead our learnable matrix $M$ towards a permutation matrix, i.e., every row and column contains precisely a single 1 with 0s elsewhere.

Recall the penalty for matrix $M$,
\begin{equation} \label{eq:penalty}
    \begin{aligned}
    \mathcal{L}_p =& \sum_{i=1}^I [\sum_{j=1}^I |M_{i,j}| - (\sum_{j=1}^I M_{i,j}^2 )^{\frac{1}{2}} ]  \\
        &+ \sum_{j=1}^I [\sum_{i=1}^I |M_{i,j}| - (\sum_{i=1}^I M_{i,j}^2 )^{\frac{1}{2}} ].
    \end{aligned}
\end{equation}
By the Cauchy-Schwarz inequality, we have
\begin{gather}
    (\sum_{j=1}^I |M_{i,j}|) - (\sum_{j=1}^I M_{i,j}^2)^{\frac{1}{2}} \geq 0, \forall i,
\end{gather}
with the equality holds if and only if there exists maximum one 1 for each row,
\begin{gather}
    |\{j:M_{i,j} \neq 0\} <= 1|, \forall i.
\end{gather}
Then, in conjunction with our normalization that perserves,
\begin{gather}
    M_{i,j} \geq 0, \forall i,j ; \\
    \sum^I_{j=1} M_{i,j} = 1, \forall i ;\sum^I_{i=1} M_{i,j} = 1, \forall j,
\end{gather}
the equality only holds if and only if there exists only one 1 for each row,
\begin{gather}
    |\{j:M_{i,j} \neq 0\} = 1|, \forall i.
\end{gather}
Likewise,
\begin{gather}
    (\sum_{i=1}^I |M_{i,j}|) - (\sum_{i=1}^I M_{i,j}^2)^{\frac{1}{2}} \geq 0, \forall j,
\end{gather}
with equality if and only if $|\{j:M_{i,j} \neq 0\} = 1|, \forall i$. 
Therefore, when the penalty (\ref{eq:penalty}) becomes 0, $M$ converges to a permutaion matrix.
For more detailed mathematical proof about the Lipschitz penalty, we refer readers to \citet{lyu2019autoshufflenet}.

\end{document}